# Temporal Pattern Discovery for Accurate Sepsis Diagnosis in ICU Patients

Eitam Sheetrit,[1,2] Nir Nissim,[1,2] Denis Klimov,[3] Lior Fuchs,[4] Yuval Elovici,[1,2] Yuval Shahar[1,3]

[1]Department of Software and Information Systems Engineering, Ben-Gurion University of the Negev, Beer-Sheva, Israel
{eitams,yshahar,elovici}@bgu.ac.il

[2]Malware Lab, Cyber Security Research Center, Ben-Gurion University of the Negev, Beer-Sheva, Israel
nirni.n@gmail.com

[3]Medical Informatics Research Center, Ben-Gurion University of the Negev, Beer-Sheva, Israel
klimov@post.bgu.ac.il

[4]Soroka University Medical Center, Beer-Sheva, Israel
liorfuchs@gmail.com

**Abstract.** Sepsis is a condition caused by the body's overwhelming and life-threatening response to infection, which can lead to tissue damage, organ failure, and finally death. Common signs and symptoms include fever, increased heart rate, increased breathing rate, and confusion. Sepsis is difficult to predict, diagnose, and treat. Patients who develop sepsis have an increased risk of complications and death and face higher health care costs and longer hospitalization. Today, sepsis is one of the leading causes of mortality among populations in intensive care units (ICUs). In this paper, we look at the problem of early detection of sepsis by using temporal data mining. We focus on the use of knowledge-based temporal abstraction to create meaningful interval-based abstractions, and on time-interval mining to discover frequent interval-based patterns. We used 2,560 cases derived from the MIMIC-III database. We found that the distribution of the temporal patterns whose frequency is above 10% discovered in the records of septic patients during the last 6 and 12 hours before onset of sepsis is significantly different from that distribution within a similar period, during an equivalent time window during hospitalization, in the records of non-septic patients. This discovery is encouraging for the purpose of performing an early diagnosis of sepsis using the discovered patterns as constructed features.

## 1 Introduction

Sepsis is a life-threatening condition that arises when the body's response to infection injures its own tissues and organs [1]. It may cause a complex variety of physiological, pathological, and biochemical abnormalities, which can lead to organ failure and death. Common signs and symptoms include fever, increased heart and breathing rate, and confusion. Patients who develop sepsis have an increased risk of complications and death, and face higher health care costs and longer hospitalization. Sepsis is a major public health concern, accounting for more than $23.6 billion (6.2%) of total US hospital costs in 2013 [2]. Today, sepsis is one of the leading causes of mortality among patients in intensive care units (ICUs), and it is responsible for 10% of the ICU admissions [3] and for the occupation of approximately 25% of the ICU beds in US hospitals [4]. Sepsis is usually treated with antibiotics and intravenous fluids. Early diagnosis is crucial to proper sepsis management. Diagnosis relies on the presence of overt symptoms of systemic illness (abnormal or rapid change in blood pressure, body temperature, heart rate, respiratory rate, etc.). After the onset of sepsis, the effectiveness of antibiotic treatment rapidly decreases.

Research shows that for each one hour delay in the administration of antibiotic treatment in a case of sepsis, the mortality rate increases by 7% [5]. Moreover, patients who survive sepsis often have long-term physical, psychological, and cognitive disabilities with significant health care and social implications [6]. Early and accurate prediction of the onset of sepsis could facil-itate effective and targeted treatment, which could, in turn, reduce the patient death rate and lower the risk of organ damage. Because of the heterogeneous nature of possible infections that can cause sepsis and varying patient response to treatment, sepsis is difficult for physicians to recognize.

Recently proposed detection methods suffer from low performance rates [7], fail to utilize all of the relevant patient data available [8], and often require laboratory tests with results obtained after a relatively long period of time [9]. With the rapid improvement in hospital information systems, an increasing amount of data about each patient admitted for treatment is being collected and stored. Much of these data are associated with concrete time stamps known as temporal data - data that varies over time. They denote the evolution of an object's characteristic over a period of time. For instance, decreasing patient blood pressure in the last hour, increasing heart rate for 30 minutes, high body temperature for 3 days, etc. In addition, with the emerging trend of extensive use of machine learning and data mining methods, which mine and leverage informative features from rough data (including temporal data), many challenging tasks in biomedical informatics have been already solved, including the detection and classification of severe conditions [10].

In our work, we propose a new approach for helping physicians predict and detect sepsis at an early stage, using the data regularly collected on each patient in the ICU. To this end, we look at the problem from a temporal data mining perspective and look for distributions of temporal patterns that might differentiate patients who eventually become septic from those who do not. To evaluate our proposed approach, we have applied it to the MIMIC-III dataset [11], limited to ICU patients who are 15 years old or older. Our evaluation shows that our approach has the potential of being beneficial to physicians in the classification or prediction of sepsis in ICU patients: the different distribution of temporal patterns in each class of patients suggests that these patterns might serve as effective features for several possible machine-learning classification methods, as had been shown in the past in several other clinical domains [12].

The rest of this paper is organized as follows. The next section provides background on sepsis and temporal data mining. Related work is discussed in section 3. The methods used in our study are described in section 4. Section 5 presents the experimental evaluation, and the results obtained based on our proposed approach are described in section 6. Section 7 concludes the paper.

## 2   Background

In this section, we provide a clinical definition of sepsis and discuss its diagnosis and introduce the field of temporal data mining including abstractions methods and the mining process itself.

### 2.1   Sepsis Definition

A redefinition of sepsis has recently been introduced in order to increase accuracy in the identification of septic patients [1]. It defines sepsis as life-threatening organ dysfunction caused by a dysregulated host response to infection. Organ dysfunction can be identified as an increase in the sequential organ failure assessment (SOFA) score [13] of at least two points. Other scoring systems currently used for the determination of sepsis include the systemic inflammatory response syndrome (SIRS), and the newly proposed bedside scoring system called qSOFA (for "quick SOFA"). SIRS is nonspecific and can be caused by ischemia, inflammation, trauma, infection, or a combination of several insults. The idea behind SIRS was to define a clinical response to a nonspecific insult of infectious or noninfectious origin. SIRS requires two or more of the following: temperature >38°C or <36°C, heart rate >90/min, respiratory rate >20/min or $Paco_2$ <32 mm Hg, WBC (white blood cell) count >12,000/mm$^3$ or <4,000/mm$^3$ or >10% immature bands. qSOFA is a new measure, intended to provide simple bedside criteria to identify adult patients with suspected infection who are likely to have poor outcomes [1]. Its criteria are: respiratory rate ≥22/min, systolic blood pressure ≤100 mm Hg, and altered menta-tion (Glasgow Coma Scale score of less than 15). Although qSOFA is less robust than SOFA in the ICU, it does not require laboratory tests and can be assessed quickly and repeatedly. All of these methods generate risk scores according to various patient vital signs and laboratory test results. However, they ignore changes and trends in patient data, and temporal relationships between different measurements.

## 2.2 Temporal Data Mining

The field of temporal data mining is concerned with the analysis and extraction of hidden predictive information, and identifying temporal patterns and regularities in time-oriented temporal data, such as sensor readings, stock market data, network monitoring, and patient data stored in medical records. In temporal data mining the entity (e.g., a sensor, computer, patient) is sampled constantly at fixed or varying time periods. The data may be represented as time points for some variables (e.g., patient's heart rate of 65/min on 01/03/17 at 17:50:00) or as time intervals (e.g., patient's antibiotic treatment from 01/03/17 until 10/03/17). Analyzing time-oriented data enables researchers to discover new knowledge in the form of temporal patterns hiding in the data and gain understanding regarding temporal behavior and associations, with the aim of supporting various classification and prediction tasks [14]. The varying forms of temporal data and different sampling times present a challenge in the analysis stage. In order to discover knowledge, the different variables need to be similarly represented. To this end, we propose using not the raw, time-stamped data, but rather an *interval-based temporal abstraction* of the patient's data.

### 2.2.1 Temporal Abstractions

Temporal abstraction (TA) is the task of representing timestamped raw data as a set of time intervals, often at a higher level of abstraction [15]. Given a set of timestamped data, external events, and abstraction goals, TA produces abstractions of the data that represent past and present states and trends that are relevant for the given set of goals. Temporal abstraction techniques usually include a discretization module as a pre-processing step, which discretizes raw data into a small number of possible values. The discretization module can be automatic [16, 17, 18, 19] or use knowledge acquired from a domain expert, a technique also known as *knowledge-based temporal abstraction* (KBTA) [20]. KBTA is often useful in medical domains, in which significant domain-experts' knowledge exists. Temporal abstraction, which in most cases includes some form of interpolation [21], may be helpful when dealing with data that was sampled at different frequencies or is missing values, or when trying to mine time-points and time-intervals together, due to its smoothing effect on the generated abstractions.

Figure 1 describes an example of a temporal abstraction of a patient's body temperature measurements over time. The time points were abstracted in this particular case into three States ("low" [less than 36°C], "normal" and "high" [higher than 38°C]) and into three Gradients (Decreasing, Stable and Increasing). Context-sensitive temporal-interpolation functions are used to create the appropriate intervals [21].

### 2.2.2 Knowledge-Based Temporal Abstraction

Introduced by Shahar [20], knowledge-based temporal abstraction (KBTA) is a multi-step method used to create abstract, interval-based concepts at multiple levels of abstraction, from timestamped raw data, by exploiting domain specific knowledge. Thus, the cut-off values for determining states, such as low body temperature might be suggested in a context-sensitive fashion by a domain expert. The KBTA output includes several types of temporal abstractions: (1) a state abstraction takes one or more values as input and generates the value of the corresponding condition (e.g., low body temperature) as output, (2) a gradient abstraction defines an interval during which the value of a parameter changes (e.g., increasing body temperature), and (3) a rate abstraction summarize the rate of change (e.g., rapid increase of body temperature).

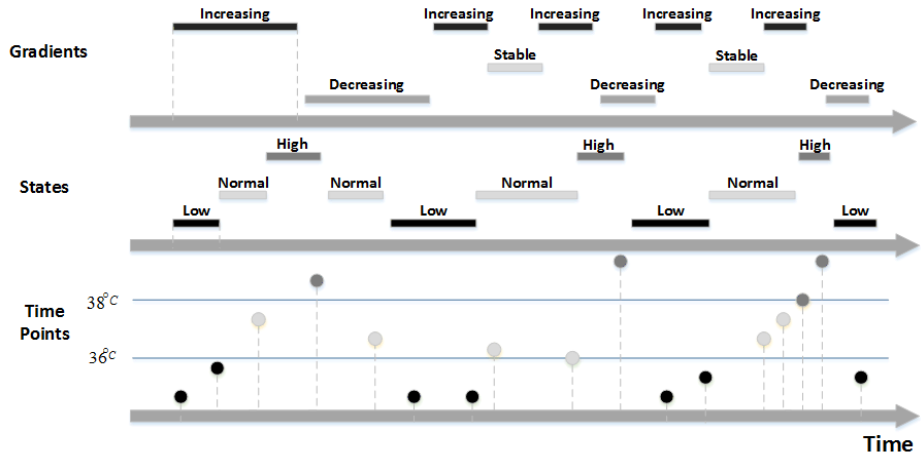

**Fig. 1.** A series of raw time-point data for a patient's body temperature is presented at the bottom. The data in this case are abstracted according to their values into three interval-based states (in the middle), and into gradient abstractions (sign of the first derivative).

Other outputs types include interval-based interpretation contexts [22], as well as various types of temporal patterns, such as linear or repeating (periodic) temporal patterns [23, 24].

### 2.2.3 Mining Time-Intervals
Time-interval mining is the task of finding frequent relations between time-intervals that exist in the time-oriented data. Most time-interval mining methods rely on Allen's temporal relations [25] to represent the temporal logic between time intervals. In [26] Hoppner introduced a method to mine rules in symbolic interval sequences using Allen's temporal relations. In this work, he defined a non-ambiguous representation of time-interval patterns by a matrix in order to represent all of the pairwise relations within a k-intervals pattern. Papapetrou et al. [27] presented a time interval mining method that corresponds to Hoppner's definition, using a hybrid approach, which indexes all of the pairs of time intervals and then extend them in candidate generation fashion. They only used five temporal relations: meets, matches (same as equal in Allen's relations), overlaps, contains, and follows, and introduced an epsilon threshold to make the temporal relations more flexible. IEMiner (Patel et al. [28]) is another method used to efficiently discover frequent temporal patterns from interval-based data, using a lossless hierarchical representation of patterns, blacklists of windows not containing enough frequent events, and dominant events (events with the latest time in the pattern) for optimized candidate pattern generation. Moskovitch and Shahar proposed an even more efficient method, called the KarmaLego framework [29]), which exploits the transitivity property of temporal relations to generate candidates efficiently. They have also shown that the discovered patterns can be used for classification and prediction purposes [12].

### 2.2.4 KarmaLego
KarmaLego [29] is a fast time-intervals mining algorithm that enumerates all of the *time-interval relations patterns* (TIRPs) whose frequency is above a given support threshold. It is based on a flexible version of Allen's seven relations, and limits the *"before"* relation by a maximal allowed gap, as proposed by Winarko and Roddick in [30]. The use of only seven basic Allen relations is possible, because the time interval pattern collection can be sorted lexicographically.

## 3    Related Work
In several papers, researchers have tried to tackle the problem of septic shock prediction, but few have attempted to predict sepsis. Septic shock is a subset of sepsis in which underlying circulatory and cellular abnormalities are profound enough to substantially increase mortality [1]. Since septic shock is an aggravation of sepsis, detection of sepsis at an early stage is preferred, and will prevent the patient from deteriorating into a septic shock, a serious life-threating condition. Most of the proposed methods have been built and tested using the MIMIC database. Desautels et al [8] proposed a sepsis predicting method based on a machine learning classification system, called InSight, that uses multivariable combinations of easily obtained patient data (vitals, peripheral capillary oxygen saturation, Glasgow Coma Score, and age), to predict sepsis using the MIMIC-III dataset, limited to ICU patients 15 years or older, who stayed in the ICU for at least seven hours prior getting sepsis. The features used by InSight are the changes in the values of each of these clinical types of data, between the point we wish to make a prediction, and two hours back. Nonlinear function approximations for each posterior probability of sepsis (s=1) given a smoothed estimate of a single clinical variable are created for the prediction score. Following the Sepsis-3 definition of the sepsis syndrome [1], these authors compared the classification performance of InSight, qSOFA, SIRS, and SOFA, assessing their ability to predict whether or not patients would become septic. In testing, InSight performed as well as or better than the traditional scoring systems for data collected 1-2 hours prior to the patient reaching a sepsis state, with an average F-score of 0.48. However, it should be noted that since the MIMIC-III database does not include the precise time in which a patient became septic, it is in theory possible that predicting sepsis very close to the sepsis onset time might be affected by measurements performed during or after the patient was diagnosed as septic.

Other recent work has focused on early prediction of septic shock. A septic shock early warning system (EWS) was developed using multivariate logistic regression on commonly measured clinical variables [9]. Using a dataset with 65 septic shock and 185 sepsis only patients, the model could predict the onset of septic shock one hour in advance with an AUC of 0.928. However, the system used invasively-gathered data and extracted features from the MIMIC waveform data, which provide higher time resolution than data more commonly available in most ICUs.

Another study performed a recursive partitioning and regression tree (RPART) analysis on 1864 septic patients to identify early predictors from the clinical data of hospitalized non-ICU patients [7]. The model, which required results from 11 routine laboratory tests and basic vital signs, correctly identified 55% of the septic shock patients.

To conclude, predicting the relatively rare event of septic shock using accessible clinical data typically did not result in very accurate models. Previous research did not usually try to focus on the problem of predicting sepsis.

## 4      Methods

This research is aimed at enhancing the ability to diagnose and detect (or even predict) sepsis in ICU patients by using the discovery of temporal interval-based patterns within the patients' clinical data. This research is aimed at developing an approach for the early diagnosis of sepsis in ICU patients, based on the application of temporal abstraction and time-interval pattern discovery on patients' clinical data.

**Sepsis Knowledge-base and Temporal Abstraction**. The first step when dealing with time-oriented raw data is creating a meaningful interpretation and representation of the data. In this work, we used the KBTA method [20], exploiting domain-expert knowledge on sepsis. We identified 26 clinical concepts related to sepsis [1], which can be divided into two groups: (1) lab tests (including laboratory measurements for patients); concepts that belong to this group include albumin, bilirubin, chloride, creatinine, fibrinogen, glucose, hemoglobin, lactate, PCO2, PH, phosphate, platelets, PO2, sodium, TCO2, urea, and white blood cell count, (2) chart items (including all charted data for patients); concepts that belong to this group are blood pressure (systolic, diastolic, mean, and pulse pressure), heart rate, minute ventilation and respiratory rate, body temperature, and Glasgow Coma Scale (GCS). Each concept was ab-stracted into three possible states (Low, Normal, High), and three possible gradient values (Decreasing, Increasing, Stable) based on its raw values and using the temporal-abstraction knowledge provided by our sepsis domain expert. The result of this step is a representation of the data as temporal intervals describing different concepts with a corresponding state or gradient, following the start and end time, as illustrated in Figure 1.

**KarmaLego**. The next step is the discovery of temporal patterns hiding in the interval collection. To this end, we use KarmaLego [29], as mentioned in section 2.2.4. We search for *time-interval relations patterns* (TIRPs) whose frequency is above 10% support threshold, in every class of patients (septic and non-septic) separately, in order to capture the unique temporal behavior of each class, which allows us to characterize and differentiate these two classes from each other. We then conducted statistical analysis on two levels: global, with a two-sample Kolmogorov-Smirnov test [31], and local, with proportion test [32]. The Two-sample Kolmogorov–Smirnov test quantifies a distance between the empirical distribution functions of two samples. The null distribution of this statistic is calculated under the null hypothesis that the samples are drawn from the same distribution. This test will allow us to investigate whether the two distributions of TIRPs discovered in the two classes are different. While Kolmogorov-smirnov captures the whole set of TIRPs and may be affected by TIRPs appearing in only one class, the proportion test can test a single TIRP at a time. While using the support value of each TIRP (the proportion of the distinct entities having the TIRP in their data records within the whole population), we can perform a test for significance of difference between a TIRP's proportions in the two classes.

## 5      Evaluation

Our evaluation mainly focused on the question of whether using knowledge-driven temporal abstraction and the discovery of frequent time-interval patterns within multivariate raw data collected in ICUs can help physicians identify or predict sepsis at an early stage. In order to do so, we pose the following research questions:

1. Is the discovered TIRPs distribution within the two classes [sepsis/non-sepsis] different?
2. If so, which patterns are the best indicators for early diagnosis of sepsis in ICU patients, and when do they appear?

**Dataset**. Our evaluation uses the MIMIC-III (Medical Information Mart for Intensive Care III) dataset [11]. It is a large, freely-available database comprised of de-identified health-related data associated with over 40,000 patients who stayed in critical care units at Beth Israel Deaconess Medical Center between 2001 and 2012. The database includes information such as demographics, vital sign measurements made at the bedside (about one data point per hour), laboratory test results, procedures, medications, caregiver notes, imaging reports, and mortality (both in and out of hospital). It consists of 26 tables that record every piece of data for the patient's admission to the hospital. With the help of our domain expert [LF], we identified 26 clinical concepts related to sepsis and then established knowledge-based state and gradient abstraction definitions. The definitions were used for the abstraction of the raw data into more meaningful concepts. Table 1 describes the knowledge base used for the process of abstraction. The gray colored columns describe laboratory blood tests, and the white colored columns describe vital signs and other easily assessed bedside measurements. Each concept is abstracted into three states: Normal (the value mentioned in the table), Low (values that are lower than normal), and High (values higher than normal); and three gradients: Stable (a change in value that is equal to or smaller than the one mentioned in the table), Increasing (a positive change larger than the value appearing in the table), and Decreasing (a negative change with an absolute value larger than the value appearing in the table).

   **Experimental Plan**. Because the precise sepsis onset time is not included in the MIMIC-III database, the first task in our experiment was to identify patients with sepsis and determine its onset time. Patients with sepsis were identified using the ICD-9 (International Classification of

Diseases, Ninth Revision) diagnosis codes (995.91 and 995.92), if they appear in a patient's discharge forms, and whether they received antibiotics during their stay in the ICU. The sepsis onset time was determined as the earliest between being prescribed antibiotics, and the time the qSofa criteria indicated the existence of sepsis. We examined the data of adult patients (15 years old or older), that stayed in the ICU for at least four days and, in the case of those that developed sepsis, only those patients who developed sepsis on the third day of their stay or later, in order to be able to collect enough data prior sepsis onset. Those conditions resulted in a total of 2,560 patients: 1,041 septic and 1,519 non-septic patients. Table 2 presents the similar age and gender distributions of patients in both groups.

**Table 1.** the Sepsis related knowledge base. Laboratory blood tests appears in gray background. vital signs and bedside measurements appears without a background color.

| Clinical Concept | "Normal" State Abstraction | Gradient Abstraction | Clinical Concept | "Normal" State Abstraction | Gradient Abstraction |
|---|---|---|---|---|---|
| Albumin | 3.4-5.4 g/dL | $\Delta > 0.5$ | Urea | 10 – 20 mg/dL | $\Delta > 5$ |
| Bilirubin | 0.2 - 1.2 mg/dL | $\Delta > 0.5$ | Sodium | 135 – 145 mEq/L | $\Delta > 5$ |
| Chloride | 96 – 106 mEq/L | $\Delta > 5$ | TCO2 | 22 – 28 mmol/l | $\Delta > 2$ |
| Creatinine | 0.6-1.3 mg/dL | $\Delta > 0.2$ | WBC | 4.5 – 10 x$10^9$/L | $\Delta > 1$ |
| Fibrinoge | 200 – 400 mg/dL | $\Delta > 50$ | Body Temperature | 36 – 38 °C | $\Delta > 0.5$ |
| Glucose | 70 – 100 mg/dL | $\Delta > 10$ | Glasgow Coma Scale* | 8 - 12 | $\Delta > 2$ |
| Hemoglobin | 11 – 18 g/dL | $\Delta > 2$ | Diastolic Blood Pressure | 70 – 90 mmHg | $\Delta > 10$ |
| Lactate | 0.5 - 2.2 mmol/L | $\Delta > 1$ | Systolic Blood Pressure | 110 – 140 mmHg | $\Delta > 10$ |
| PCO2 | 38 – 42 mm Hg | $\Delta > 2$ | Mean Blood Pressure | 65 - 80 | $\Delta > 5$ |
| PH | 7.34 - 7.45 pH | $\Delta > 0.05$ | Heart-Rate | 60 – 80 bpm | $\Delta > 10$ |
| Phosphate | 2.4 - 4.1 mg/dL | $\Delta > 0.5$ | Minute-Ventilation | 5.4 – 11 L/min | $\Delta > 0.5$ |
| PLT | 150 – 400 x$10^9$/L | $\Delta > 50$ | Pulse-Pressure | 35 – 45 mmHg | $\Delta > 5$ |
| PO2 | 75 – 100 torr | $\Delta > 10$ | Respiratory-Rate | 7 – 14 breath/pm | $\Delta > 3$ |

\* Glasgow Coma Scale states are mild, moderate and severe

**Table 2.** Age and gender distributions for patients of the two groups

| Test Description | % Male/Female | Age <30 | 30< Age <50 | 50< Age <70 | 70 < Age |
|---|---|---|---|---|---|
| Septic group | 61% / 39% | 5% | 14% | 39% | 42% |
| Non-septic group | 58% / 42% | 4% | 17% | 37% | 42% |

*Experiment 1*: The purpose of this experiment was to examine the behavior of Septic ICU patients in the last 12 hours prior to sepsis onset, compared to non-septic ICU patients in equivalent 12 hours. As most of the patients (about 75%) became septic on their 4[th] day in ICU, we chose to look at the 4[th] day's last 12 hours for the non-septic patients as the equivalent control-group time. In order to do so, we extracted the identified 26 sepsis related clinical concepts for each patient with their temporal timestamp. While vital signs and bedside measurements in the MIMIC-III database are being recorded every hour (and sometimes even every 15 minutes), laboratory blood tests are less frequent, and being performed only once a day. In order to handle the low frequency, we decided that laboratory tests are gathered starting from the ICU admission time, and the other measurements are gathered only in the 12 hours window. Both gathering processes end at the end of the 12 hours window. We computed the state and gradient abstractions and the frequent interval-based temporal patterns within each class (septic and non-septic patients) separately. We then examined discovered patterns in each class that appeared (at least once) within at least 10% of the patient records belonging to that class. We used an Information-Gain feature selection method to evaluate each pattern, and select the most informative ones. Finally, we performed the two-sample Kolmogorov-Smirnov test [33] to investigate whether the distributions of discovered TIRPs in the two classes are significantly different, and proportion test for each discovered TIRP: (1) between the two classes on TIRPs discovered on both classes (2) in the same class (randomly divided into two groups) on TIRPs discovered on this class.

*Experiment 2*: This experiment was designed similarly to Experiment 1, but used temporal concepts from the only the last six hours before sepsis onset. The purpose of this experiment was to examine whether sampling the data that are closer in time to the sepsis onset increases the differences between the distributions of temporal patterns in the septic class and the non-septic class, and, as a result, discover more informative patterns according to Information-Gain feature selection methods.

## 6 Results

*Experiment 1*: In this experiment, we discovered within both classes a total of 26,968 interval-based temporal patterns that appeared at least once within at least 10% of the patients of at least one of the classes. Out of these patterns, 6,168 where exclusively found within the Septic-patients class, 6,416 where exclusively found within the non-septic class, and the rest (14,384) appeared within the longitudinal records of both patient classes. The top rated temporal patterns, according to Information-Gain feature selection method are shown in Fig. 2. Note that, on the two right most columns, we have also indicated the support level of each temporal pattern in regarding to each class (the septic and non-septic). Using the Two-sample Kolmogorov-Smirnov test, we discovered that the distribution of temporal patterns for the two classes is statistically different with a confidence level of 0.95 ($\alpha$ = 0.05), where the maximum difference (D) = 0.038 and the Critical D = 0.001. The proportion tests (also with $\alpha$ = 0.05) support the results, by presenting a significant difference in the relative amount of different proportions of TIRPs between the two classes, in compare to the relative amount within the classes, as presented in Table 3.

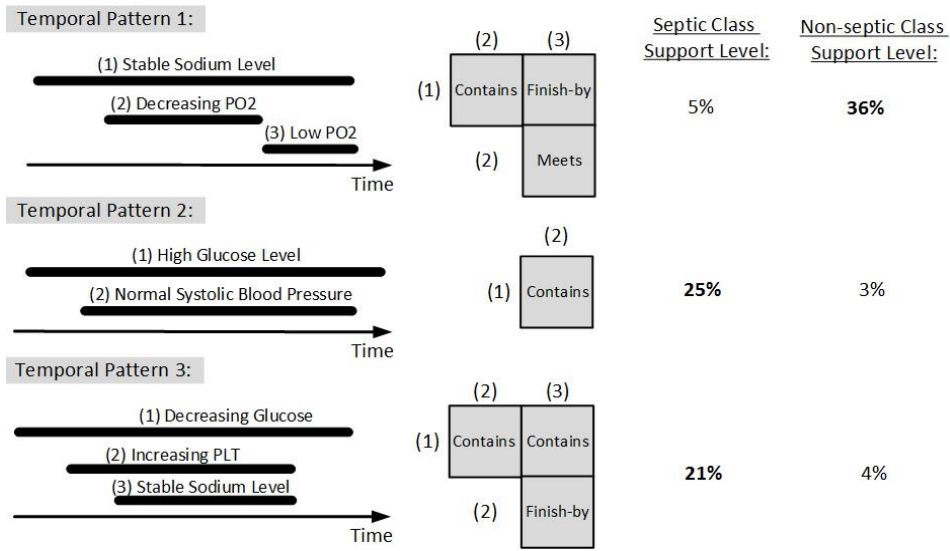

**Fig. 2.** Experiment 1's top rated temporal patterns, according to the Information-Gain feature selection methods. The level of support for each pattern appears for each of the two classes (septic and non-septic patients).

**Table 3.** Proportion test results on each of the three cases, for data collected in the last 12 hours.

| Test Description | temporal patterns amount | Different proportion amount | % difference |
| --- | --- | --- | --- |
| Septic vs. Non-septic | 14,383 | 5937 | 41% |
| Only Septic (50% vs. 50%) | 20,551 | 984 | 5% |
| Only Non-septic (50% vs. 50%) | 20,799 | 1277 | 6% |

***Experiment 2***: In the second experiment, we discovered for both classes a total of 22,422 temporal patterns that appear in at least 10% of the patients of at least one of the classes. Out of these patterns, 5,356 where exclusively found in the Septic-class, 5,349 where exclusively found on the non-septic class, and the rest (11,717) appeared within both classes. Using the Two-sample Kolmogorov-Smirnov test, we discovered that the distribution of temporal patterns within the two classes is also statistically different with a confidence level of 0.95 ($\alpha = 0.05$), where the maximum difference (D) = 0.047 and the Critical D = 0.001, and with similar results for the proportion tests (Table 4).

**Table 4.** Proportion test results on each of the three cases, for data collected in the last 6 hours.

| Test Description | temporal patterns amount | Different proportion amount | % difference |
| --- | --- | --- | --- |
| Septic vs. Non-septic | 11,716 | 5122 | 44% |
| Only Septic (50% vs. 50%) | 17,072 | 750 | 4% |
| Only Non-septic (50% vs. 50%) | 17,065 | 1076 | 6% |

# 7 Discussion and Conclusions

We have shown that the distribution of frequent (at least 10% of the patients) interval-based temporal-abstraction patterns is quite different in the longitudinal records of patients who will eventually be classified as septic or non-septic. This difference was demonstrated both within the 12 hours and the six hours periods before onset of sepsis (as determined according to our definition). This difference can be exploited in multiple ways that we are currently exploring: Using the temporal patterns as a special type of classification and prediction features within a series of machine-learning algorithms, as done by Moskovitch and Shahar [12] and by others; using clustering methods that include both temporal and non-temporal data; using KNN-like case-based reasoning methods in which the discovered TIRPs are a part of the distance function, etc. In all of these cases, the varying TIRP distributions found within the longitudinal records of septic and non-septic patients strongly encourage the potential for inducing effective classification and prediction models for the existence or the eventual appearance of sepsis in ICU patients.